\documentclass[lettersize,journal]{IEEEtran}
\usepackage{amsmath,amsfonts}
\usepackage{algorithmic}
\usepackage{algorithm}
\usepackage{array}
\usepackage[caption=false,font=normalsize,labelfont=sf,textfont=sf]{subfig}
\usepackage{textcomp}
\usepackage{stfloats}
\usepackage{url}
\usepackage{verbatim}
\usepackage{graphicx}
\usepackage{cite}
\usepackage{multirow}
\usepackage{enumitem}
\usepackage{booktabs}
\usepackage{makecell}
\begin{document}

\title{OccluDex: Hierarchical 3D Visuo--Tactile Representation Learning for Egocentric Dexterous Manipulation under Self-Occlusion}


\author{
Ziheng Xu$^{1}$,
Yueyuan Chen$^{1}$,
Xinyuan He$^{1}$,
Guoxing Liu$^{1}$,
Yuanshuo Tan$^{1}$,
Huiming Pan$^{1}$,
Bin He$^{2}$,
Shuo Jiang$^{2,\dagger}$,
Peter B. Shull$^{3,\dagger}$\\[3pt]
$^{1}$Shanghai Jiao Tong University \quad
$^{2}$Tongji University \quad
$^{3}$The Chinese University of Hong Kong, Shenzhen\\
$^\dagger$ Corresponding authors.}


\maketitle

\begin{abstract}
Reliable dexterous manipulation requires continuous estimation of object geometry and hand--object contact throughout interaction. With egocentric sensing, however, the manipulating hand frequently occludes task-relevant object surfaces and contact regions, reducing the visual evidence available for state estimation and thereby making robust closed-loop control and generalization to unseen object geometries particularly challenging. To address this, we present OccluDex, a hierarchical 3D visuo--tactile representation learning framework that integrates global geometric structure with local contact information for robust manipulation under dynamic self-occlusion during hand--object interaction. OccluDex adopts multi-scale masked autoencoding to progressively encode partial 3D geometry and fuses tactile contact tokens with high-level geometric features through cross-modal attention. The encoder is pretrained from synchronized human visuo--tactile demonstrations and transferred as a frozen perceptual backbone for downstream reinforcement learning. We evaluate OccluDex on a faucet rotation task, requiring one full clockwise handle revolution, and a tabletop object reorientation task, requiring a 180\textdegree{} tabletop object reorientation without toppling. In simulation experiments, OccluDex demonstrated 12.6\% higher accuracy for unseen objects and 8.3\% higher accuracy for previously seen objects than the strongest state-of-the-art baseline models. Physical experiments were further performed with a Shadow Hand to demonstrate successful zero-shot sim-to-real generalization on unseen physical objects. This results could enable humanoid egocentric object manipulation for seen and unseen objects even when the manipulating robotic hand occludes vision.

\end{abstract}


\section{Introduction}
\IEEEPARstart{D}{exterous} manipulation is a fundamental capability of embodied intelligence, requiring perception and control to be tightly coupled with physical interaction. Egocentric sensing provides a natural paradigm for such systems by reducing reliance on external calibration and maintaining observations in a reference frame aligned with the manipulating body\cite{hu2025egocentric}. This self-contained perception--action setting is particularly desirable for scalable deployment in unstructured environments\cite{howard2019evolving}\cite{ortenzi2019robotic}.

Dexterous manipulation typically occurs within the robot's peripersonal space~\cite{ladavas2004visuo}, where close-range egocentric sensing is inherently susceptible to hand--object self-occlusion. The hand can obscure a substantial portion of the object, particularly around the contact interface, while finger articulation and object motion continually alter the visible geometry. Consequently, the policy must operate on observations in which the target object is partially occluded by the hand and the hand--object interaction evolves over time, making robust closed-loop control particularly challenging.

\begin{figure}[t]
    \includegraphics[width=0.99\linewidth]{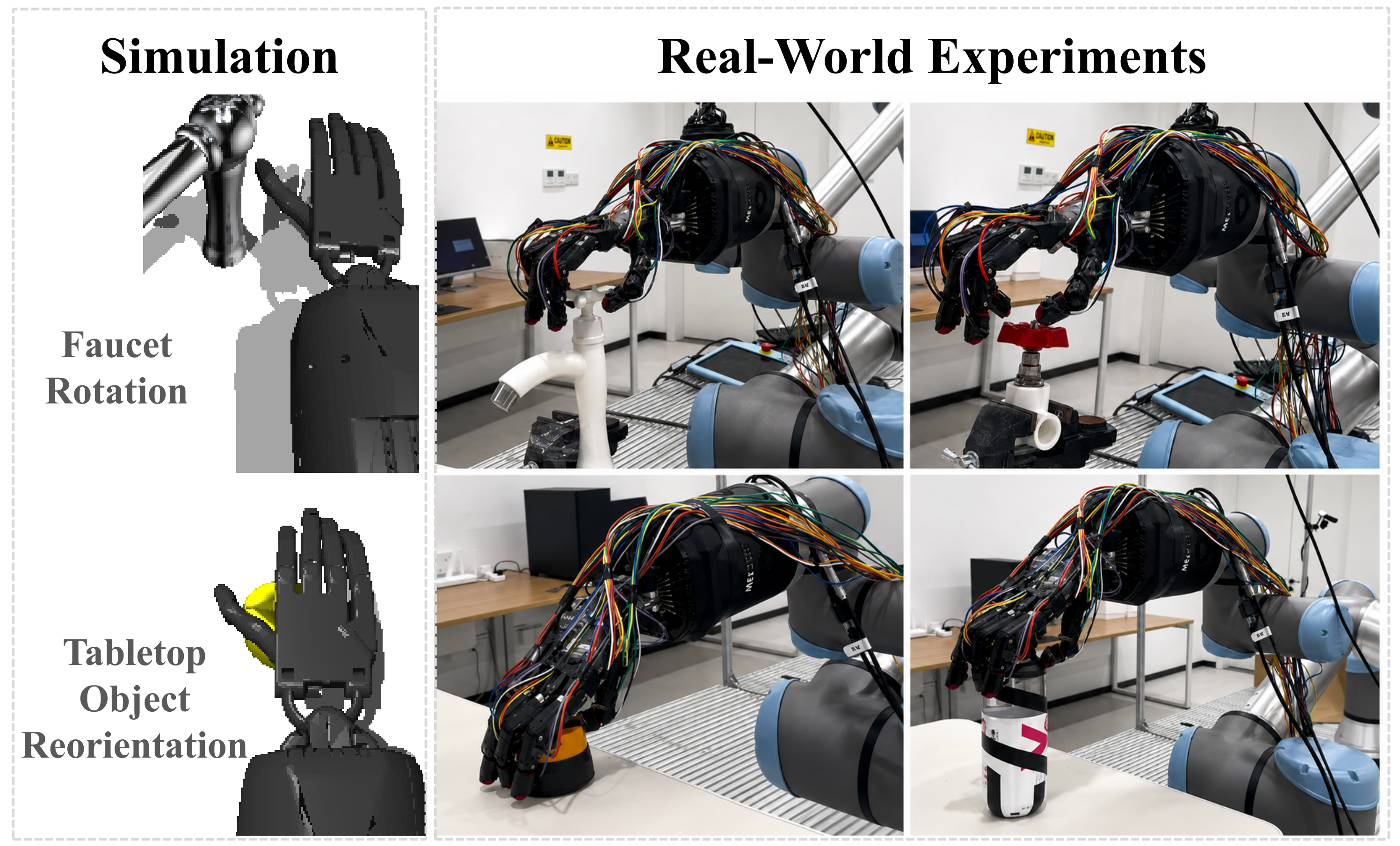}
    \caption{Simulation and real-world evaluation of OccluDex. In simulation (left), Faucet Rotation uses three faucet geometries for seen evaluation and two unseen scales for generalization testing, while Tabletop Object Reorientation uses ten seen and six unseen YCB objects. In real-world experiments (right), two previously unseen physical objects are evaluated for each task.}
    \label{fig:real_world_tasks}
\end{figure}

Recent advances in 3D completion and generative reconstruction provide powerful tools for recovering missing object geometry from partial observations \cite{yu2021pointr}\cite{huang2025compc}. These methods can substantially improve geometric completeness and have shown strong capability in inferring plausible unobserved structure. However, under severe egocentric self-occlusion, the available visual evidence may be limited and continuously changing, increasing ambiguity in geometric inference and placing additional demands on real-time reconstruction. \cite{rustler2022active}\cite{wang2025touch2shape}. More importantly, geometric completion alone does not directly provide the instantaneous contact information required for dexterous control. A key challenge is therefore to learn a control-relevant perceptual representation directly from egocentric observations under dynamic self-occlusion during hand--object interaction, without relying on explicit 3D reconstruction.

Tactile sensing provides complementary information by directly measuring physical contact. However, tactile observations are inherently local, sparse, and contact-dependent, and therefore provide limited information about global object geometry \cite{li2020review}. 
Therefore, recent studies have explored visuo--tactile integration to combine 2D visual and contact information for fine-grained manipulation \cite{chen2022visuotactile}\cite{liu2025vtdexmanip}. However, dexterous manipulation policy depends strongly on metric geometry and local surface structure, which can be explicitly represented in 3D. Explicit 3D representations provide a natural geometric reference for grounding tactile cues under severe occlusion~\cite{cai2025geometry}. Yet effectively integrating the two modalities remains nontrivial, as global geometric context and sparse local contacts differ substantially in spatial support and representation scale.

Neurophysiological studies suggest that humans mitigate similar perceptual limitations through the joint encoding of 3D visual and tactile signals by body-centered multimodal neurons with spatially aligned receptive fields, supporting a coherent representation of peripersonal space.~\cite{holmes2004body}. Such integration combines global 3D geometric structure with local contact information to support robust control under dynamic self-occlusion during hand--object interaction.

Inspired by this mechanism, we propose OccluDex, a hierarchical 3D visuo--tactile representation learning framework for dexterous manipulation under severe self-occlusion. OccluDex progressively encodes point clouds through multi-scale tokenization, masking, and hierarchical Transformer encoding to capture local-to-global geometric structure. Tactile observations are embedded as contact tokens and fused with high-level geometric features through cross-modal attention. A hierarchical decoder reconstructs masked geometric and tactile observations for joint visuo--tactile pre-training. The pretrained encoder is frozen and transferred as the perceptual backbone of a downstream proximal policy optimization (PPO) policy.

To the best of our knowledge, OccluDex is the first hierarchical 3D visuo--tactile representation learning framework for dexterous manipulation in occlusion-rich peripersonal space. Our main contributions are:
\begin{itemize}
    \item We collect an egocentric human manipulation dataset with synchronized partial 3D point clouds and tactile observations, providing paired geometric--contact supervision under natural hand--object occlusion.

    \item We introduce a hierarchical 3D visuo--tactile masked pre-training framework that integrates multi-scale geometry with tactile contact cues through cross-modal attention, enabling the learned representation to preserve both global spatial context and local interaction information under hand--object occlusion.

    \item We demonstrate improved dexterous manipulation performance and generalization across seen and unseen settings on two tasks, Faucet Rotation and Tabletop Object Reorientation, under egocentric self-occlusion, together with zero-shot sim-to-real transfer to a physical Shadow Dexterous Hand with previously unseen objects (Fig.~\ref{fig:real_world_tasks}).
\end{itemize}

\section{Related Work}

\subsection{Egocentric Perception for Dexterous Manipulation}
Egocentric perception has emerged as a promising paradigm for dexterous manipulation\cite{ze2025generalizable}\cite{kareer2025egomimic}. However, during in-hand manipulation in peripersonal space, the hand frequently occludes substantial portions of the object. Consequently, geometric information, particularly around contact regions, may become incomplete or ambiguous in visual observations \cite{qin2022dexmv}\cite{wang2025multi}\cite{pmlr-v270-huang25e}\cite{LinP-RSS-25}.

Existing approaches often mitigate visual occlusion through external viewpoints \cite{qin2022dexmv}\cite{xu2023unidexgrasp}\cite{huang2025fungrasp}\cite{openai2019rubiks} or geometric reconstruction from partial observations\cite{yu2021pointr}\cite{huang2025compc}\cite{li2025genpc}\cite{wu2026tosc}. Although effective in reducing visual ambiguity, these strategies still depend on additional visual evidence or reconstructed geometry. In peripersonal manipulation, however, extensive hand--object self-occlusion can substantially limit the available visual evidence, while the occlusion pattern changes continuously with the interaction rather than remaining static. A key challenge is therefore to learn a control-relevant perceptual representation directly from egocentric observations under dynamic self-occlusion during hand--object interaction, without relying on explicit 3D reconstruction.

\subsection{Visuo-Tactile Representation Learning}
Tactile sensing directly reflects local hand--object interaction and is particularly valuable when visual observations are incomplete, occluded, or ambiguous\cite{LinP-RSS-25}\cite{lee2024dextouch}\cite{yan2024avita}. However, tactile sensing remains inherently sparse and local, providing limited information about the global geometric structure of the manipulated object.

Recent efforts have therefore explored visuo--tactile representation learning to combine geometric perception with physical interaction. By integrating visual observations with tactile feedback, these methods provide richer representations of manipulation states and contact information\cite{chen2022visuotactile}\cite{liu2025vtdexmanip}\cite{zhu2026touch}\cite{yuan2024robot}.

Despite this progress, existing visuo--tactile representation learning methods are predominantly built upon image-based visual representations, in which spatial structure is encoded only implicitly. For dexterous manipulation under self-occlusion, 3D representations provide more explicit geometric organization of the hand, object, and contact regions, offering a stronger basis for grounding local tactile cues\cite{cai2025geometry}. Existing 3D visuo--tactile approaches have also been primarily studied with parallel-jaw grippers\cite{pmlr-v270-huang25e}, whose contact patterns and interaction dynamics are substantially simpler than those of multi-fingered dexterous hands. Consequently, structured 3D visuo--tactile representation learning for multi-fingered dexterous manipulation remains underexplored.

\section{Methods}

\begin{figure*}[!t]
\centering
\includegraphics[width=\textwidth]{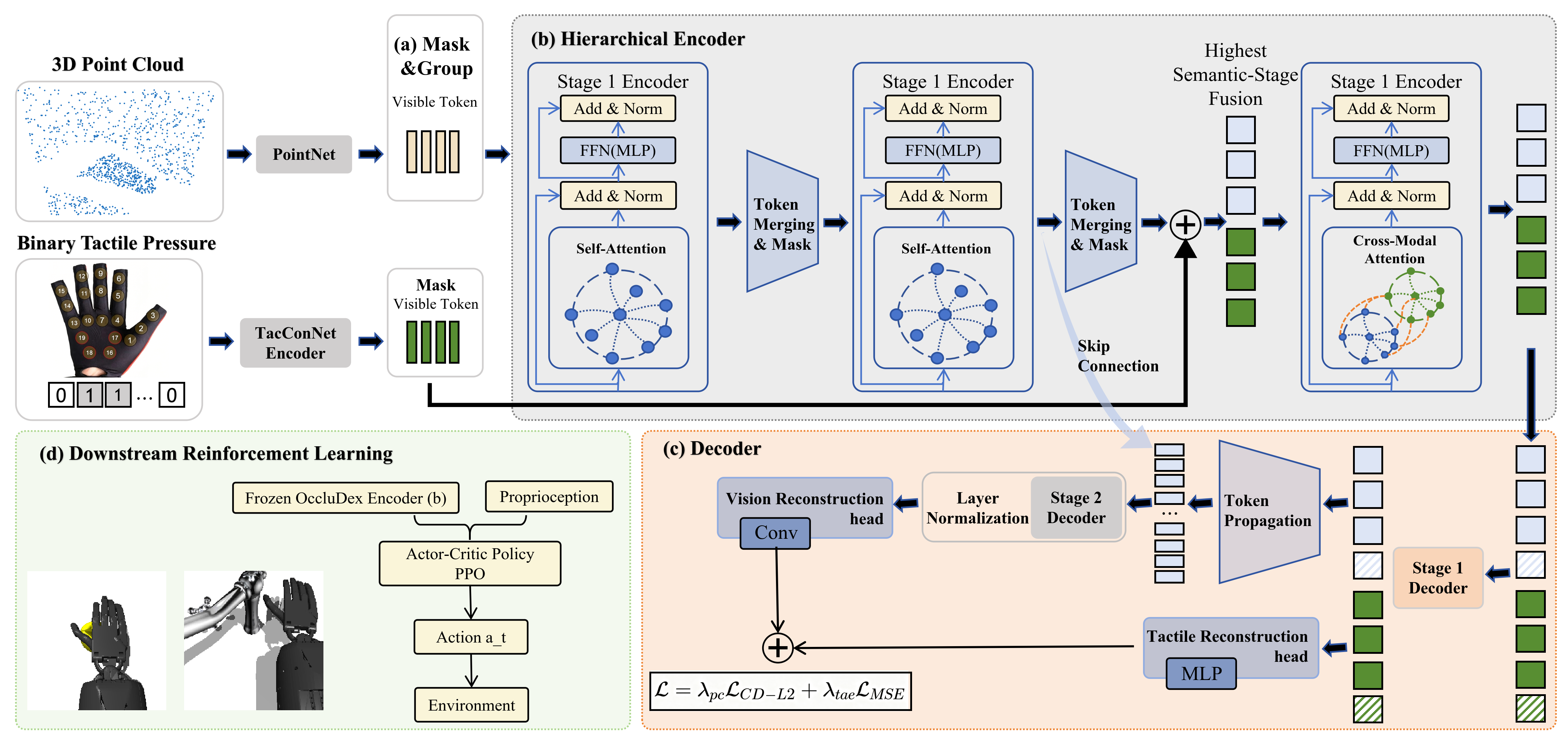}
\caption{Overview of the proposed 3D visuo--tactile representation learning framework. (a) Multi-scale grouping and masking. (b) A hierarchical encoder progressively abstracts 3D geometry and performs high-level cross-modal interaction. (c) The decoder reconstructs masked geometric and tactile observations through hierarchical token propagation. (d) The pretrained encoder is frozen and transferred to downstream PPO-based dexterous manipulation.}
\label{fig:system}
\end{figure*}

OccluDex is a hierarchical 3D visuo--tactile representation learning framework designed to extract control-relevant information from egocentric observations subject to dynamic hand--object self-occlusion (Fig.~\ref{fig:system}). Given an egocentric point cloud and synchronized tactile measurements, the geometric branch progressively aggregates local surface structures into higher-level spatial representations, while tactile measurements are encoded as sensor-specific contact tokens. The two modalities are fused through cross-modal attention at the highest semantic stage, allowing sparse contact cues to interact with global 3D geometric context. During pre-training, geometric and tactile observations are jointly masked and reconstructed through a hierarchical decoder, encouraging the encoder to learn joint geometric--tactile representations and cross-modal associations. After pre-training, the decoder is discarded and the frozen encoder is transferred as the perceptual backbone of a downstream PPO policy.
\subsection{Hierarchical Multi-scale Geometric Encoding}
Dexterous manipulation requires geometric representations that are simultaneously sensitive to fine-grained local structure and expressive of broader spatial relationships. These complementary requirements naturally span multiple spatial scales and are difficult to capture with a single-resolution representation. We therefore construct a hierarchical geometric representation that progressively aggregates local point neighborhoods into coarser semantic regions, allowing the encoder to capture fine surface structure together with broader environment context. Given an input point cloud $P_0=P$, we recursively construct a hierarchy of multi-scale geometric regions. At scale $i$, $1 \leq i \leq S$, representative geometric centers are sampled from the preceding scale using Farthest Point Sampling (FPS), followed by k-Nearest Neighbor (k-NN) grouping:
\begin{equation}
P_i = \operatorname{FPS}(P_{i-1},N_i), \qquad
I_i = \operatorname{kNN}(P_i,P_{i-1},k_i),
\end{equation}
where $P_i \in \mathbb{R}^{N_i \times 3}$ denotes the sampled centers, and $I_i \in \mathbb{N}^{N_i \times k_i}$ stores the indices of the $k_i$ nearest neighbors in $P_{i-1}$ for each center in $P_i$.

\begin{figure*}[!t]
\centering
\includegraphics[width=0.9\textwidth]{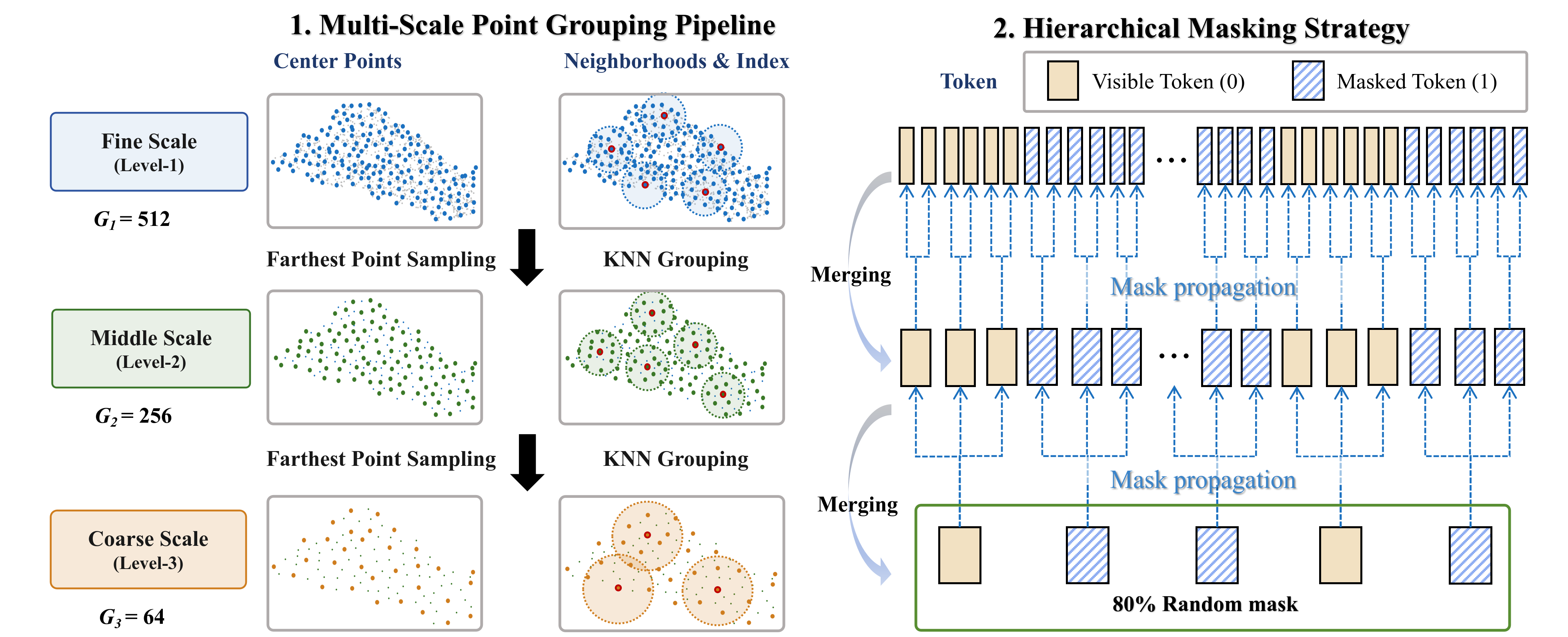}
\caption{Hierarchical multi-scale grouping and progressive masking. Multi-scale geometric regions are constructed through recursive sampling and neighborhood grouping, while the visibility pattern is propagated from the highest semantic scale to finer scales through hierarchical correspondences, preserving structural consistency across the geometric hierarchy.}
\label{fig:multiscale_masking}
\end{figure*}

Because the same physical region is represented at multiple levels of the hierarchy, independently masking each scale would create inconsistent visibility patterns across resolutions. We therefore adopt progressive multi-scale masking\cite{zhang2022point}, in which visibility is sampled at the coarsest scale and propagated toward finer scales through hierarchical neighborhood correspondences (Fig.~\ref{fig:multiscale_masking}). Let $\mathcal{V}_i$ denote the index set of visible tokens at scale $i$. Given the visible set $\mathcal{V}_S$ sampled at the prescribed geometric masking ratio, the visibility pattern is recursively propagated as
\begin{equation}
\mathcal{V}_i =
\bigcup_{j\in\mathcal{V}_{i+1}}
I_{i+1}[j,:],
\qquad i=S-1,\ldots,1,
\end{equation}
where $I_{i+1}[j,:]$ denotes the finer-scale neighborhood in $P_i$ associated with the $j$-th center in $P_{i+1}$. The corresponding masked index set is defined as
$\mathcal{M}_i=\{1,\ldots,N_i\}\setminus\mathcal{V}_i$,
and the visible and masked positions are obtained as
$P_i^v=P_i[\mathcal{V}_i]$ and $P_i^m=P_i[\mathcal{M}_i]$, respectively. The masking strategy used during pre-training is not intended to simulate the physical self-occlusion pattern itself; rather, it provides a self-supervised objective that encourages the encoder to learn joint representations of geometric observations and tactile signals and capture their cross-modal dependencies.

At the first encoder stage, visible local regions are mapped into point tokens using a lightweight PointNet-style tokenizer. For each center in $P_1^v$, the points in its neighborhood specified by $I_1$ are jointly encoded through shared point-wise transformations and neighborhood aggregation, yielding the initial token set $T_1^v \in \mathbb{R}^{N_1^v \times C_1}$.

For subsequent encoder stages, $1<i\leq S$, token merging progressively coarsens the representation. The visible tokens from the $(i-1)$-th stage are gathered into local token groups according to the hierarchical neighborhood indices $I_i$. Each group is then aggregated by combining token-wise features with pooled neighborhood context, producing the merged tokens $T_i^v \in \mathbb{R}^{N_i^v \times C_i}$. Because the visibility pattern is propagated consistently across scales, the merged tokens remain aligned with the same visible geometric regions throughout the hierarchy. As the hierarchy deepens, the number of tokens decreases while the feature dimensionality increases, allowing larger spatial regions to be represented with richer contextual information.

At each scale, the visible tokens are augmented with geometric positional embeddings and processed by the corresponding Transformer encoder.

\subsection{Tactile Token Embedding}

In parallel with geometric encoding, tactile observations are encoded into token-level representations that preserve localized contact information. Given the 19-channel tactile measurement $T \in \mathbb{R}^{19}$, each channel is binarized according to a fixed activation threshold. This binary representation emphasizes contact activation while reducing sensitivity to variations in force magnitude.

Each tactile channel is treated as an individual token and mapped into a $C$-dimensional latent representation:
\begin{equation}
Z_{tac}^{(j)}
=
\phi_{tac}(\hat{T}_j)
+
p_{tac}^{(j)}
+
e_{tac},
\qquad j=1,\ldots,19,
\end{equation}
where $\phi_{tac}$ denotes the tactile embedding function, $p_{tac}^{(j)}$ is the positional embedding associated with the $j$-th sensing location, and $e_{tac}$ is a learnable tactile modality embedding. The resulting token sequence $Z_{tac}\in\mathbb{R}^{19\times C}$ preserves sensor-specific contact information for subsequent visuo--tactile interaction.

\subsection{Cross-modal Visuo--Tactile Fusion}

During pre-training, tactile tokens are independently masked following the masked autoencoder paradigm, and only visible tactile tokens are passed to the encoder. 

We perform visuo--tactile fusion only at the highest semantic encoder stage. A central challenge is that the two modalities describe hand--object interaction at different spatial scales: geometric tokens progressively encode spatial context from local surfaces to larger regions, whereas tactile tokens represent sparse contact events at specific sensing locations. Direct fusion at shallow geometric stages would therefore combine representations with strongly mismatched spatial support. We instead delay cross-modal interaction until the highest semantic stage, where the geometric hierarchy has aggregated sufficient global context to provide a stable spatial reference for tactile cues. The final-stage geometric and tactile tokens are concatenated and processed by the cross-modal Transformer. Their corresponding positional embeddings are concatenated in the same manner, while learnable modality embeddings are used to distinguish geometric and tactile tokens.

The joint sequence is subsequently processed by the final Transformer encoder through cross-modal attention. To exclude invalid padded geometric entries while preserving cross-modal interactions, we construct a block-wise attention mask
\begin{equation}
M_{\text{joint}} =
\begin{bmatrix}
M_{pc\text{-}pc} & M_{pc\text{-}tac} \\
M_{tac\text{-}pc} & M_{tac\text{-}tac}
\end{bmatrix},
\end{equation}
where $M_{pc\text{-}pc}$ preserves the geometric attention mask, $M_{pc\text{-}tac}$ and $M_{tac\text{-}pc}$ suppress interactions involving invalid geometric tokens, and $M_{tac\text{-}tac}$ permits interactions among visible tactile tokens.
\subsection{Hierarchical Decoder and Reconstruction}

Starting from the coarsest geometric scale, masked geometric and tactile observations are reconstructed using a lightweight hierarchical decoder. Learnable mask tokens are inserted at masked positions and combined with the encoded visible tokens to form the decoder input. The decoder consists of $S-1$ lightweight Transformer stages, where global self-attention facilitates information exchange between visible and masked tokens during reconstruction.

Tactile reconstruction is performed after the first decoder stage. Since tactile observations do not possess an explicit multi-scale spatial hierarchy, the decoded tactile features are directly mapped to the masked tactile predictions through a lightweight reconstruction head.

Geometric decoding, in contrast, proceeds hierarchically toward finer spatial resolutions. Between subsequent decoder stages, token propagation interpolates coarse latent features onto finer geometric centers and fuses them with the corresponding encoder features through skip connections. The decoder progressively recovers fine-grained geometric information while retaining features learned during hierarchical encoding.

The decoded geometric representations are mapped to the masked local point patches through a lightweight reconstruction head. We optimize geometric reconstruction using the symmetric Chamfer Distance:
\begin{equation}
\mathcal{L}_{pc}
=
\frac{1}{|\hat{P}_m|}
\sum_{\hat{p}\in\hat{P}_m}
\min_{p\in P_m}
\|\hat{p}-p\|_2^2
+
\frac{1}{|P_m|}
\sum_{p\in P_m}
\min_{\hat{p}\in\hat{P}_m}
\|p-\hat{p}\|_2^2,
\end{equation}
where $P_m$ denotes the ground-truth masked point patches.

For tactile reconstruction, the loss is computed only over the masked tactile tokens:
\begin{equation}
\mathcal{L}_{tac}
=
\frac{1}{|\mathcal{M}_t|}
\sum_{j \in \mathcal{M}_t}
\left\|
\tilde{t}_j-t_j
\right\|_2^2,
\end{equation}
where $\mathcal{M}_t$ denotes the index set of masked tactile tokens.

The overall pre-training objective is defined as
\begin{equation}
\mathcal{L}
=
\lambda_{pc}\mathcal{L}_{pc}
+
\lambda_{tac}\mathcal{L}_{tac},
\end{equation}
where $\lambda_{tac}$ balances the geometric and tactile reconstruction objectives. Joint masked reconstruction encourages the encoder to learn complementary representations that capture both 3D geometric structure and interaction-relevant contact information.

\section{Experiments}

We evaluate OccluDex in terms of representation effectiveness, generalization, and real-world transfer on downstream dexterous manipulation tasks. We first describe the experimental setup in Section~\ref{sec:exp_setup}, followed by the evaluation metrics and baselines in Section~\ref{sec:metrics_baselines}. Quantitative comparisons with representative visuo--tactile representation learning methods are presented in Section~\ref{sec:comparison}, while Section~\ref{sec:ablation} analyzes the contribution of the proposed pre-training components through ablation studies. Finally, Section~\ref{sec:real_world} evaluates sim-to-real transfer on a Shadow Dexterous Hand using previously unseen objects.

\subsection{Experimental Setup}
\label{sec:exp_setup}

\textbf{Human Demonstration Data Collection.}
We collect an egocentric visuo--tactile demonstration dataset for representation pre-training using a head-mounted Intel RealSense D435i camera and a WiseGlove equipped with 19 tactile sensing channels. The camera captures RGB-D observations of hand--object interactions from the operator's viewpoint, while the glove records contact responses during manipulation. The visual and tactile streams are synchronized at 30~Hz to form paired visuo--tactile observations.

Data are collected from three human subjects performing six representative manipulation tasks: Upright Object Reorientation, Tabletop Object Reorientation, Object Pickup, Bottle-Cap Manipulation, Object Sliding, and Screw Manipulation. The three subjects contribute 565, 565, and 560 sequences, respectively, resulting in 1,690 sequences and 409,035 synchronized visuo--tactile frames in total.

\textbf{Simulation Setup.}
We pretrain the representation on two NVIDIA GeForce RTX 4080 GPUs, with the main hyperparameters summarized in Table~\ref{tab:pretrain_setting}. The pretrained representation is evaluated on downstream reinforcement learning tasks in NVIDIA Isaac Gym using a simulated Shadow Dexterous Hand. We consider two representative tasks: Faucet Rotation and Tabletop Object Reorientation \cite{liu2025vtdexmanip}. Both tasks involve substantial hand-induced self-occlusion during manipulation, with the objective of stably rotating the faucet handle or object through one full revolution. Unless otherwise specified, the reward functions and other simulation parameters follow \cite{liu2025vtdexmanip}. At each control step, the policy receives a 48-D proprioceptive state, a 19-D binary tactile observation, and a partial point cloud of 4,096 points. The egocentric camera resolution is set to $424\times240$, and the simulated tactile activation threshold is 0.01. Both tasks operate at 60~Hz. Task-specific environment configurations are summarized in Table~\ref{tab:env_setting}. We place the camera in an egocentric configuration, where the hand naturally occludes a substantial portion of the target object during manipulation (Fig.~\ref{fig:real_world_tasks})

\begin{table}[t]
\centering
\caption{Main hyperparameters used for OccluDex pre-training.}
\label{tab:pretrain_setting}
\begin{tabular}{lc}
\hline
\textbf{Parameter} & \textbf{Value} \\
\hline
Tactile threshold & 40 \\
Point-cloud mask ratio & 0.8 \\
Tactile mask ratio & 0.5 \\
Group sizes & $[16,\,8,\,8]$ \\
Number of groups & $[512,\,256,\,64]$ \\
Encoder dimensions & $[96,\,192,\,384]$ \\
Encoder depths & $[2,\,2,\,2]$ \\
Decoder depths & $[1,\,1]$ \\
Effective batch size & 128 \\
Point-cloud loss weight & 50 \\
\hline
\end{tabular}
\end{table}

\begin{table}[t]
\centering
\caption{Environment configurations for the downstream manipulation tasks.}
\label{tab:env_setting}
\begin{tabular}{lcc}
\hline
\textbf{Parameter} & \makecell{\textbf{Tabletop Object}\\\textbf{Reorientation}} & \textbf{Faucet Rotation} \\
\hline
Parallel environments & 200 & 150 \\
Episode length & 600 & 500 \\
Camera position & $(0,\,0.25,\,0.5)$ & $(0.25,\,0,\,0.5)$ \\
Camera target & $(0,\,-0.1,\,-0.2)$ & $(-0.1,\,0,\,-0.2)$ \\
\hline
\end{tabular}
\end{table}

\subsection{Metrics and Baselines}
\label{sec:metrics_baselines}
\textbf{Metrics.}
We use success rate as the primary evaluation metric for all downstream dexterous manipulation tasks. For each random seed, the trained policy is evaluated over 100 episodes in both seen and unseen settings, and we report the mean and standard deviation across three seeds.

\textbf{Baselines.}
We compare OccluDex against representative state-of-the-art visuo--tactile representation learning methods that explicitly exploit cross-modal interactions between visual and tactile observations for downstream dexterous manipulation. \textbf{VTDexManip}~\cite{liu2025vtdexmanip} learns transferable 2D visuo--tactile representations from human manipulation data and applies the learned representation to downstream reinforcement learning. \textbf{VTT-3D} is adapted from the original VTT formulation to accommodate 3D point-cloud observations \cite{chen2022visuotactile}. We make only the minimal modifications required for the 3D setting while preserving its core cross-modal visuo--tactile learning objective and network design, and denote the resulting variant as VTT-3D. \textbf{Base-only} uses only proprioceptive observations as policy input. All methods are evaluated under the same downstream PPO training protocol, training budget, and evaluation procedure.

\subsection{Comparison on Downstream Dexterous Manipulation}
\label{sec:comparison}

For Faucet Rotation, three faucet instances are used during training, while unseen generalization is evaluated by scaling the faucet geometries by factors of 0.9 and 1.1. For Tabletop Object Reorientation, ten objects from the YCB dataset \cite{calli2015ycb} are used for training, and six held-out YCB objects are reserved for unseen-object evaluation.

\begin{figure}[t]
    \centering
    \includegraphics[width=\linewidth]{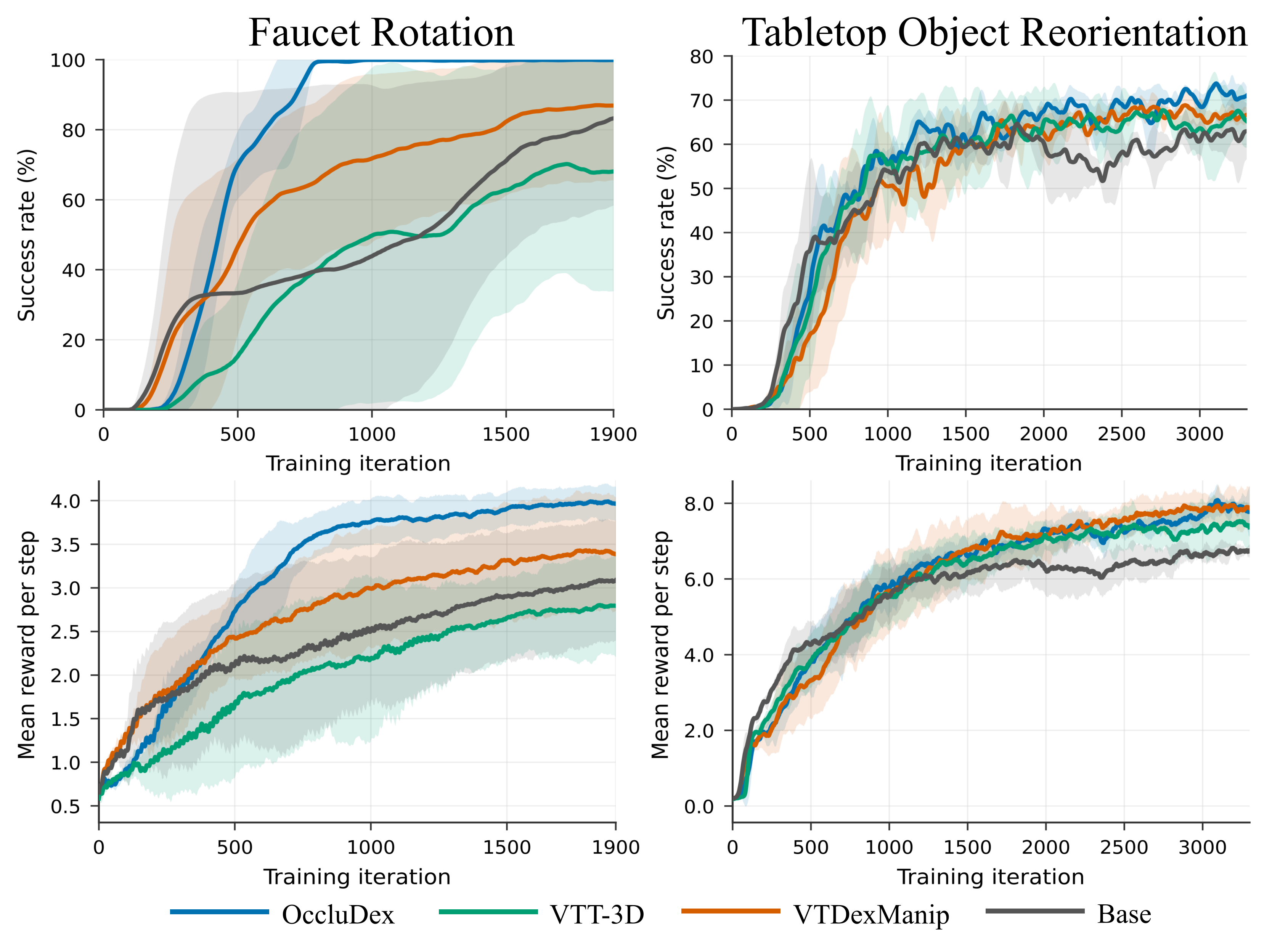}
    \caption{Training curves on Faucet Rotation and Tabletop Object Reorientation. The top row shows success rate, and the bottom row shows mean reward per step. Solid lines denote the mean over three random seeds, and shaded regions indicate the corresponding standard deviation.}
    \label{fig:downstream_training}
\end{figure}

The learned representation facilitates downstream policy optimization, as reflected in the training dynamics on both downstream tasks (Fig.~\ref{fig:downstream_training}).

Table~\ref{tab:downstream_success} summarizes the evaluation results. On Faucet Rotation, OccluDex achieves 99.9\% success on seen instances and 86.6\% on unseen faucet scales, outperforming all baselines in both settings. Its advantage is particularly evident on unseen scales. OccluDex also exhibits substantially lower cross-seed variability on this task, indicating more consistent performance across independently trained policies. Averaged over seen and unseen evaluations, OccluDex reaches 93.2\% success.

On Tabletop Object Reorientation, OccluDex again achieves the highest success rates, reaching 75.1\% on seen objects and 83.3\% on held-out objects. Its average success rate of 79.2\% exceeds VTDexManip at 76.9\%, VTT-3D at 72.1\%, and Base at 63.9\%, demonstrating effective generalization to previously unseen objects.

Overall, OccluDex achieves an average success rate of 86.2\% across the two downstream tasks, compared with 75.8\% for VTDexManip, 73.0\% for Base, and 65.6\% for VTT-3D. Together with the training dynamics (Fig.~\ref{fig:downstream_training}), these results demonstrate consistent gains across distinct dexterous manipulation tasks and evaluation settings.

\begin{table*}[t]
\centering
\caption{Evaluation results on Faucet Rotation and Reorient Object on Table. Values are mean $\pm$ standard deviation over three random seeds.}
\label{tab:downstream_success}
\small
\setlength{\tabcolsep}{4.5pt}
\renewcommand{\arraystretch}{1.15}
\begin{tabular}{llcccccc}
\hline
\multirow{2}{*}{\textbf{Task}} &
\multirow{2}{*}{\textbf{Method}} &
\multicolumn{3}{c}{\textbf{Seen}} &
\multicolumn{3}{c}{\textbf{Unseen}} \\
\cline{3-5} \cline{6-8}
&
&
\textbf{Success (\%)} &
\textbf{Return} &
\textbf{Ep. Len.} &
\textbf{Success (\%)} &
\textbf{Return} &
\textbf{Ep. Len.} \\
\hline

\multirow{4}{*}{Faucet Rotation}
& Base-only
& $83.8 \pm 23.9$
& $842.8 \pm 168.9$
& $292.0 \pm 120.3$
& $80.6 \pm 14.9$
& $810.4 \pm 129.3$
& $318.6 \pm 106.5$ \\

& VTDexManip
& $86.3 \pm 20.5$
& $743.3 \pm 110.1$
& $238.3 \pm 96.6$
& $63.0 \pm 25.6$
& $714.9 \pm 56.6$
& $339.5 \pm 85.9$ \\

& VTT-3D
& $69.7 \pm 29.5$
& $896.5 \pm 133.6$
& $340.5 \pm 112.9$
& $48.8 \pm 26.9$
& $813.1 \pm 52.7$
& $404.8 \pm 77.1$ \\

& \textbf{OccluDex}
& $\mathbf{99.9 \pm 0.2}$
& $643.2 \pm 39.3$
& $\mathbf{157.9 \pm 16.9}$
& $\mathbf{86.6 \pm 2.9}$
& $636.3 \pm 25.9$
& $\mathbf{229.8 \pm 7.1}$ \\

\hline

\multirow{4}{*}{\makecell{Tabletop Object\\Reorientation}}
& Base-only
& $65.1 \pm 5.7$
& $573.2 \pm 118.9$
& $77.1 \pm 12.4$
& $62.7 \pm 1.6$
& $554.9 \pm 154.3$
& $84.2 \pm 23.6$ \\

& VTDexManip
& $72.1 \pm 1.5$
& $651.9 \pm 62.4$
& $77.3 \pm 5.4$
& $81.8 \pm 3.9$
& $611.7 \pm 17.4$
& $\mathbf{76.0 \pm 3.8}$ \\

& VTT-3D
& $68.5 \pm 3.0$
& $595.9 \pm 52.8$
& $\mathbf{75.2 \pm 7.3}$
& $75.7 \pm 6.4$
& $572.5 \pm 32.9$
& $76.4 \pm 11.8$ \\

& \textbf{OccluDex}
& $\mathbf{75.1 \pm 2.3}$
& $660.0 \pm 20.4$
& $81.0 \pm 7.5$
& $\mathbf{83.3 \pm 2.0}$
& $577.8 \pm 35.7$
& $78.8 \pm 15.4$ \\

\hline
\end{tabular}
\end{table*}

\subsection{Ablation Study}
\label{sec:ablation}
Table~\ref{tab:ablation} summarizes the ablation results on both downstream tasks, where we evaluate the contributions of tactile feedback, 3D geometric observations, and visuo--tactile pre-training to downstream manipulation performance.
\begin{enumerate}[label=\arabic*)]

\item \textbf{Contribution of tactile feedback.} Removing tactile input causes the largest degradation on both tasks. On Faucet Rotation, the point-cloud-only variant reduces success by 45.7\% on seen instances and 48.0\% on unseen scales. On Tabletop Object Reorientation, the corresponding drops are 15.6\% on seen objects and 17.3\% on held-out objects. These results highlight the critical role of tactile feedback in providing contact information unavailable from geometry alone.

\item \textbf{Contribution of geometric observations.} The tactile-only variant consistently underperforms the full model. Incorporating point-cloud observations improves success by 24.3\% and 6.7\% on the seen and unseen evaluations of Faucet Rotation, respectively, and by 9.6\% and 4.7\% on Tabletop Object Reorientation. This confirms that global geometric information complements local tactile cues and contributes to stronger downstream performance.

\item \textbf{Effect of Pre-training.} Training the full OccluDex encoder jointly with PPO from scratch incurs prohibitively high computational and memory costs, making full-scale training impractical under our hardware constraints. Reducing the number of parallel environments to five makes training feasible but results in 0\% success, indicating the practical difficulty of jointly learning perception and control at this scale rather than isolating the effect of pre-training. We therefore remove the hierarchical design and adopt a simplified single-stage fusion architecture as a feasible from-scratch reference. OccluDex consistently outperforms this reference, with larger gains on unseen and held-out evaluations. Overall, the proposed pre-training strategy offers two practical advantages: the frozen encoder substantially reduces the computational cost of downstream reinforcement learning, while its learned representation provides stronger downstream performance than the feasible non-hierarchical from-scratch reference.

\end{enumerate}

\begin{table}[t]
    \centering
    \caption{Ablation study. Values denote success rates (\%).}
    \label{tab:ablation}
    \setlength{\tabcolsep}{3.8pt}
    \renewcommand{\arraystretch}{1.12}
    \begin{tabular}{llcc}
        \toprule
        Task & Variant & Seen (\%) & Unseen (\%) \\
        \midrule
        \multirow{4}{*}{Faucet Rotation}
        & \textbf{OccluDex (full)}        & $\mathbf{100.0}$ & $\mathbf{83.7}$ \\
        & w/o point cloud                 & $75.7$           & $77.0$ \\
        & w/o tactile                     & $54.3$           & $35.7$ \\
        & From scratch (w/o hierarchy)    & $98.7$           & $76.8$ \\
        \midrule
        \multirow{4}{*}{\makecell{Tabletop Object\\Reorientation}}
        & \textbf{OccluDex (full)}        & $\mathbf{77.3}$ & $\mathbf{81.3}$ \\
        & w/o point cloud                 & $67.7$           & $76.7$ \\
        & w/o tactile                     & $61.7$           & $64.0$ \\
        & From scratch (w/o hierarchy)    & $66.4$           & $68.3$ \\
        \bottomrule
    \end{tabular}
\end{table}

\subsection{Real-World Experiments}
\label{sec:real_world}

\begin{figure}[t]
    \centering
    \includegraphics[width=0.95\linewidth]{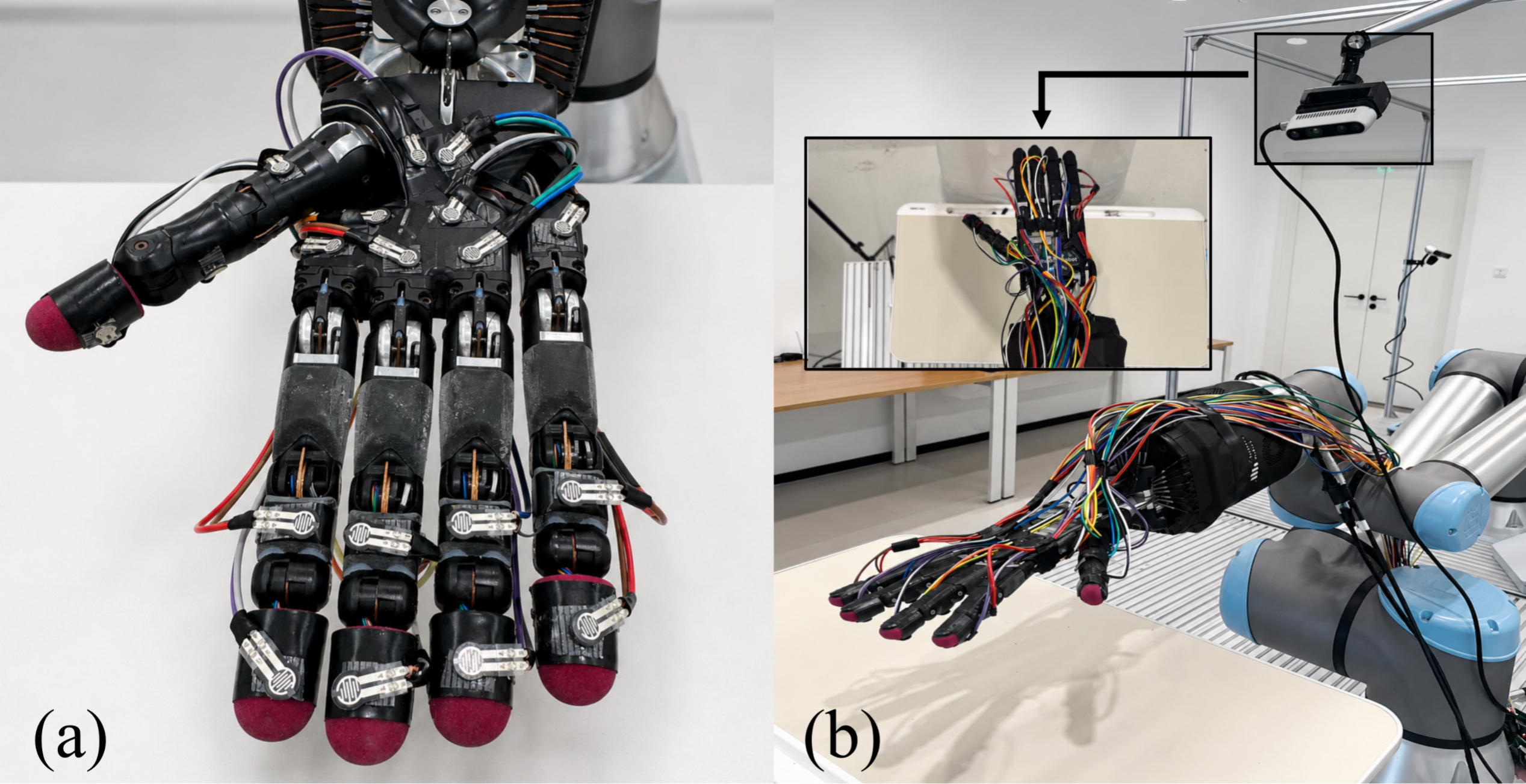}
    \caption{Real-world experimental setup. (a) Distribution of the tactile sensors on the Shadow Dexterous Hand. The sensors are attached to the fingertips, finger links, and palm to provide contact observations during manipulation. (b) Overview of the real-world manipulation workspace, including the robotic arm, the tactile-instrumented Shadow Hand, the overhead RealSense D435i camera, and the corresponding egocentric camera observation used for point-cloud perception.}
    \label{fig:real_hardware}
\end{figure}

\paragraph{Hardware setup} Our real-world platform consists of a right Shadow Dexterous Hand instrumented with 19 force-sensitive resistive sensors, an Intel RealSense D435i depth camera, and a host computer for synchronized sensing and policy execution. The tactile sensors are distributed across the fingertips, finger links, and palm (Fig.~\ref{fig:real_hardware}). Their resistance changes are converted to voltage signals through conditioning modules and sampled at 1~kHz using three USB data-acquisition (DAQ) devices. The measurements are then binarized into the same 19-dimensional contact representation used in simulation. The D435i is mounted above the workspace with a fixed view of the hand--object interaction region and provides depth observations at 30~Hz, from which a partial point cloud of 4,096 points is constructed. Together with 24 joint positions and 24 joint velocities, these observations form the multimodal input to the policy. The policy operates in closed loop at 10~Hz and outputs 20 normalized joint-position targets, which are mapped to the corresponding Shadow Hand actuators for execution.

\paragraph{Sim-to-real transfer} We perform zero-shot sim-to-real transfer without updating the pre-trained visuo--tactile encoder or PPO policy using real-world data. During simulation training, robustness is encouraged by randomizing the initial hand configuration and object pose and by adding Gaussian noise to proprioceptive observations and policy actions. Training does not involve dynamics randomization or perturbations to the point-cloud observations. The real-world camera placement is matched to the simulation setup to preserve the egocentric sensing configuration. Under this viewpoint, the Shadow Hand substantially occludes the target object during both Faucet Rotation and Tabletop Object Reorientation, resulting in persistently incomplete object geometry throughout manipulation.

\paragraph{Closed-loop validation} We further evaluate OccluDex on two real-world dexterous manipulation tasks using physical objects that are not observed during simulation training. For each task, we select two unseen objects and perform 10 trials per object, resulting in 20 trials per task. For Faucet Rotation, a trial is considered successful if the hand completes one full rotation of the faucet handle within 10~s. For Tabletop Object Reorientation, a trial is considered successful if the object is stably reoriented through at least 180\textdegree{} within 60~s. OccluDex succeeds in 17 of 20 Faucet Rotation trials and 14 of 20 Tabletop Object Reorientation trials, corresponding to success rates of 85.0\% and 70.0\%, respectively. These results demonstrate that the simulation-trained policy can be effectively executed on unseen physical objects and sustain closed-loop dexterous manipulation without real-world fine-tuning.

\section{Conclusion}
\label{sec:conclusion}
In this paper, we proposed OccluDex, a hierarchical 3D visuo--tactile representation learning framework for dexterous manipulation under severe self-occlusion. The framework integrates global 3D geometric structure with local tactile contact cues through masked pre-training and transfers the learned encoder to downstream PPO-based manipulation. Experiments on Faucet Rotation and Tabletop Object Reorientation demonstrate consistent improvements over proprioceptive and visuo--tactile baselines, with OccluDex achieving the highest average success rate of 86.2\% across the two tasks. The simulation-trained policy further transfers zero-shot to a real Shadow Dexterous Hand, achieving success rates of 85.0\% and 70.0\% on previously unseen physical objects. Ablation results further confirm the contributions of 3D geometry, tactile contact cues, and visuo--tactile pre-training. These results establish hierarchical 3D visuo--tactile representation learning as a promising perceptual foundation for robust and transferable egocentric dexterous control under dynamic self-occlusion. Future work will explore richer and higher-density tactile sensing, including lateral fingertip contacts, with particular emphasis on scalable representation learning and fusion between dense tactile observations and 3D geometry. We will further extend the framework to finer-grained dexterous manipulation tasks that require more precise contact perception and control.

\bibliographystyle{IEEEtran}
\bibliography{references}

\end{document}